\documentclass[10pt,twocolumn,letterpaper]{article}

\usepackage{cvpr}
\usepackage{times}
\usepackage{epsfig}
\usepackage{graphicx}
\usepackage{amsmath}
\usepackage{amssymb}
\usepackage{xcolor}
\usepackage{algorithm}
\usepackage{algpseudocode}
\usepackage{float}
\usepackage{cuted}
\usepackage{capt-of}
\usepackage{blindtext} 
\usepackage{lipsum} 
\usepackage{enumitem}

\usepackage[font=small,skip=3pt]{caption}
\usepackage[breaklinks=true,bookmarks=false]{hyperref}

\cvprfinalcopy % *** Uncomment this line for the final submission

\def\cvprPaperID{6215} % *** Enter the ICCV Paper ID here

\begin{document}

%%%%%%%%% TITLE
\title{Robust 3D Self-portraits in Seconds}

\author{Zhe Li$^1$, Tao Yu$^1$, Chuanyu Pan$^1$, Zerong Zheng$^1$, Yebin Liu$^{1,2}$\\
$^1$Department of Automation, Tsinghua University, China\\
$^2$Institute for Brain and Cognitive Sciences, Tsinghua University, China
}

\maketitle
%\pagestyle{fancy}
%\thispagestyle{empty}

%%%%%%%%% ABSTRACT
%\input{sections/0_abstract.tex}

%%%%%%%%% BODY TEXT
%\input{sections/1_introduction.tex}
%\input{sections/2_related_work.tex}
%\input{sections/3_overview.tex}
%\input{sections/4_method.tex}
%\input{sections/7_results.tex}
%\input{sections/8_diss.tex} 

\begin{abstract}
In this paper, we propose an efficient method for robust 3D self-portraits using a single RGBD camera. Benefiting from the proposed PIFusion and lightweight bundle adjustment algorithm, our method can generate detailed 3D self-portraits in seconds and shows the ability to handle subjects wearing extremely loose clothes.
To achieve highly efficient and robust reconstruction, we propose PIFusion, which combines learning-based 3D recovery with volumetric non-rigid fusion to generate accurate sparse partial scans of the subject. Moreover, a non-rigid volumetric deformation method is proposed to continuously refine the learned shape prior.
Finally, a lightweight bundle adjustment algorithm is proposed to guarantee that all the partial scans can not only ``loop'' with each other but also remain consistent with the selected live key observations.
The results and experiments show that the proposed method achieves more robust and efficient 3D self-portraits compared with state-of-the-art methods.
\end{abstract}

\section{Introduction}
\label{sec_intro}
Human body 3D modeling, aiming at reconstructing the dense 3D surface geometry and texture of the subject, is a hot topic in both computer vision and graphics and is of great importance in the area of body measurement, digital content creation, virtual try-ons, etc.
Traditional human body 3D modeling methods usually rely on experts for data capture and are therefore hard to use.
Compared with traditional 3D scanning methods, 3D self-portrait methods, which allow users to capture their own portraits without any assistance, have significant potential for wide usage.

Current 3D self-portrait methods can be classified into 3 categories: learning-based methods, fusion-based methods, and bundle-adjustment-based methods.
Learning-based methods mainly focus on 3D human recovery from a single RGB image (\cite{kanazawaHMR18, pifuSHNMKL19}). Thus, the results are still far from accurate due to occlusions and depth ambiguities.
Fusion-based methods reconstruct scene geometries in an incremental manner, so error accumulation is inevitable, especially for non-rigid scenarios \cite{newcombe2015dynamic}, which is detrimental for loop closure reconstruction (e.g., 3D self-portraits).
To suppress the accumulated error in incremental fusion, another branch of 3D self-portrait methods also utilizes bundle adjustment algorithms \cite{li20133d, tong2012scanning, cui2013kinectavatar, dou20153d, wang2017templateless, wang2018dynamic}. The whole sequence is first segmented into several chunks, and then fusion methods are applied to each chunk to fuse a smooth partial scan. Finally, non-rigid bundle adjustment is used to ``loop'' all the partial scans simultaneously by non-rigid registration based on explicit loop closure correspondences and bundling correspondences.
Although RGBD bundle adjustment methods have achieved state-of-the-art performance for 3D self-portraits, they still suffer from complicated hardware setups (e.g., relying on multiple sensors or electric turntables \cite{tong2012scanning,web_texel,web_shapify} or low efficiency \cite{li20133d, alldieck2018video, dou20153d, cui2013kinectavatar,  wang2017templateless, wang2018dynamic}).

\begin{figure}[t]
	\centering
	\includegraphics[width=\linewidth]{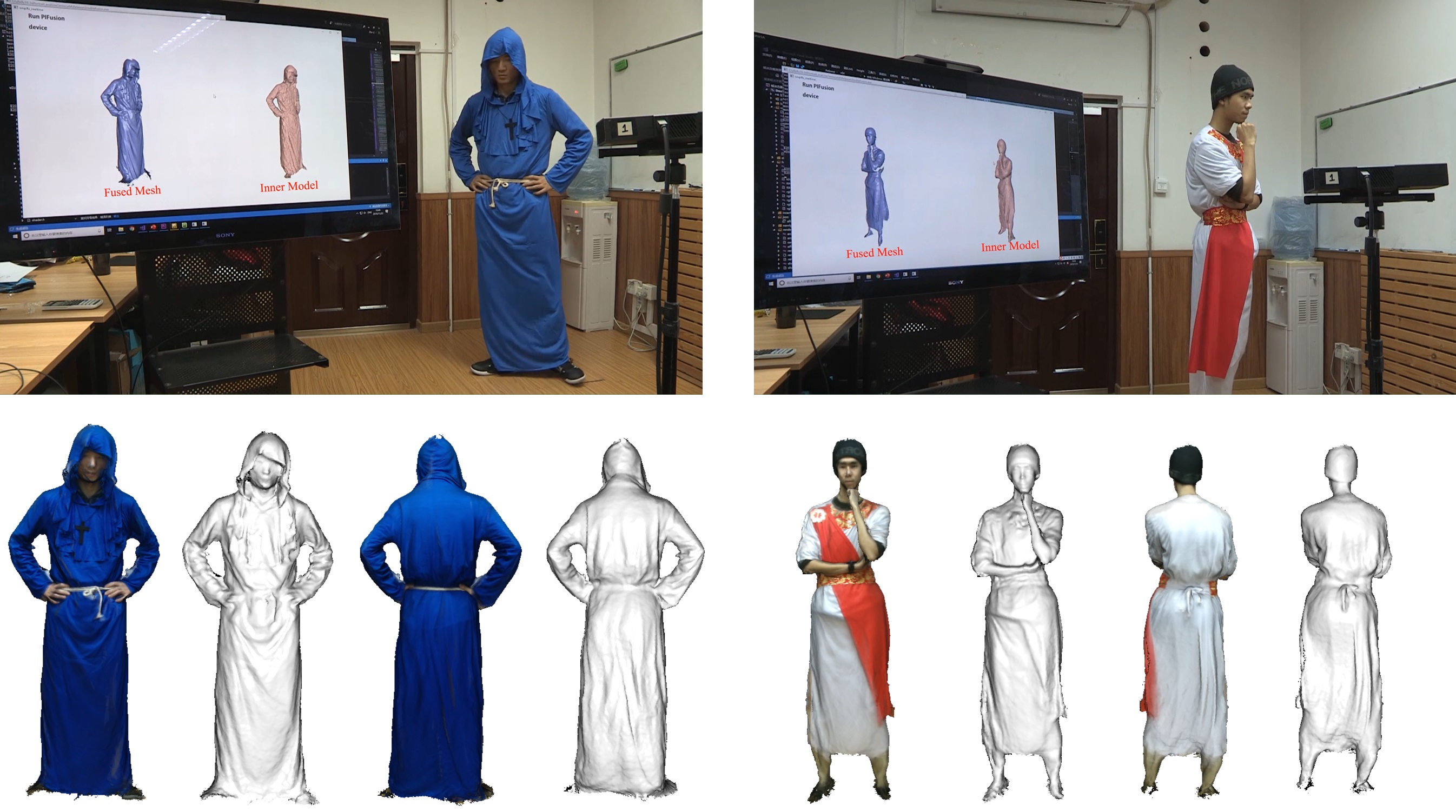}
\caption{Our system reconstructs a detailed and textured portrait after the subject self-rotates in front of an RGBD sensor.}
	\label{fig:teaser}
\end{figure}

One of our observations is that a good combination of non-rigid fusion and bundle adjustment should guarantee both efficiency and accuracy.
However, non-rigid fusion methods (e.g., \cite{newcombe2015dynamic} etc.) usually suffer from heavy drifts and error accumulation during tracking, which limit their ability to generate accurate large partial scans.
This limitation has led to the fact that previous bundle adjustment methods have to be conducted on considerably large numbers of small partial scans, which significantly increase the optimizing variables in the bundling step. For example, in \cite{dou20153d}, 40-50 small partial scans need to be bundled together, which takes approximately 5 hours.

To produce large and accurate partial scans by non-rigid fusion, a complete shape prior is necessary. To this end, we propose {PIFusion}, which utilizes learning-based 3D body recovery (PIFu \cite{pifuSHNMKL19}) as an inner layer in non-rigid fusion \cite{newcombe2015dynamic}.Specifically, in each frame, the inner layer generated by learning-based methods acts as a strong shape prior to improve the tracking accuracy and robustness, and the fused mesh in return improves the accuracy of the inner layer by the proposed non-rigid volumetric deformation (Sec.~\ref{subsec:volumetric opt}). We also improve the original PIFu \cite{pifuSHNMKL19} by incorporating pixel-aligned depth features for more accurate and robust inner-layer generation (Fig. \ref{fig:comp of pifu}).

Another important observation is that to generate accurate portraits, all the partial scans produced by PIFusion should not only construct a looped model (\cite{dou20153d, li20133d}) but also always remain consistent with the real-world observations, especially the depth point clouds and silhouettes.
Instead of using the dense bundle method in \cite{dou20153d, wang2017templateless}, we contribute a lightweight bundle adjustment method that involves live terms, key frame selection and joint optimization. Specifically, during each iteration, all the partial scans are not only optimized to ``loop''' with each other in the reference frame but are also warped to fit each key input in live frames. The key frames are selected adaptively according to the proposed live depth/silhouette energies. This method further improves the bundle accuracy without losing efficiency.

In summary, by carefully designing the reconstruction pipeline, our method integrates all the advantages from learning, fusion and bundle adjustment methods while avoiding the disadvantages and finally enables efficient and robust 3D self-portraits using a single RGBD sensor.

The contributions can be summarized as follows:
\begin{itemize}[leftmargin=*]
\setlength{\itemsep}{0pt}
\setlength{\parsep}{0pt}
\setlength{\parskip}{0pt}
	\item A new 3D self-portrait pipeline that leverages fusion, learning and bundle adjustment methods and achieves efficient and robust 3D self-portrait reconstruction using a single RGBD sensor.
	\item A new non-rigid fusion method, PIFusion, which combines a learning-based shape prior and a non-rigid volumetric deformation method to generate large and accurate partial scans.
	\item A lightweight bundle adjustment method that involves key frame selection and new live energy terms to jointly optimize the loop deformation in the reference frame, as well as the warp fields to live key frames, and finally improves the bundling accuracy without losing efficiency.
\end{itemize}

\section{Related Work}
\label{related_work}

\subsection{Learning-based 3D Human Recovery}
Learning-based 3D body reconstruction has become more and more popular in recent years. By ``seeing'' a large amount of ground truth 3D human models, current deep neural networks can infer plausible 3D bodies from various easy-to-obtained inputs, e.g., a single RGB image\cite{kanazawaHMR18, omran2018NBF,kolotouros2019cmr, Pavlakos2019CVPR, valentin2019moulding, alldieck2019tex2shape, zhu2019detailed, Natsume_2019_CVPR, pifuSHNMKL19, zheng2019deephuman}. For example, Kanazawa \textit{et al.} \cite{kanazawaHMR18}, Omran \textit{et al.} \cite{omran2018NBF} and Kolotouros \textit{et al.} \cite{kolotouros2019cmr} proposed to directly regress the parameters of a statistical body template from a single RGB image. Zhu \textit{et al.} \cite{zhu2019detailed} and Alldieck \textit{et al.} \cite{alldieck2019tex2shape} took a step forward by deforming the body template according to shading and silhouettes in order to capture more surface details. To address the challenge of varying cloth topology, recent studies have explored many 3D surface representations for deep neural networks, including voxel grids\cite{zheng2019deephuman}, multi-view silhouettes\cite{Natsume_2019_CVPR}, depth maps\cite{valentin2019moulding} and implicit functions\cite{pifuSHNMKL19}. Although these methods enable surprisingly convenient 3D human capture, they fail to generate detailed and accurate results due to occlusions and inherent depth ambiguities. 

\subsection{3D Human Using Fusion-based Methods}
In fusion-based methods, given a noisy RGBD sequence, the scene geometry is first registered to each frame and then updated based on the observations. As a result, the noise in the depth map can be significantly filtered out and the scene can be completed in a incremental manner. The pioneer work in this direction is KinectFusion\cite{Newcombe2011KinectFusion}, which was designed for rigid scene scanning using a RGBD sensor. Thus, when scanning live targets like humans, the subjects are required to keep absolutely static in order to get accurate portraits, which is not consistent with the fact that humans are ultimately moving. To handle this problem, Zeng \textit{et al.} \cite{Zeng_2013_CVPR} proposed a method for quasi-rigid fusion, but it still relies on rotating sensors for data capture, which is hard-to-use. DynamicFusion\cite{newcombe2015dynamic} extended KinectFusion and contributed the first non-rigid volumetric fusion method for real-time dynamic scene reconstruction. Following works \cite{innmann2016volume, slavcheva2017cvpr, Slavcheva_2018_CVPR, guo2017real, chao2018ArticulatedFusion, yu2017bodyfusion, zheng2018hybridfusion} kept improving the performance of DynamicFusion by incorporating different types of motion priors or appearance information. For instance, based on the double-layer surface representation, DoubleFusion\cite{yu2018doublefusion} achieved state-of-the-art performance for dynamic human body reconstruction (with implicit loop closure) using non-rigid fusion. However, constrained by the parametric inner layer representation, DoubleFusion has limited performance for reconstructing extremely wide clothes like long skirts and coats. Moreover, the A-pose requirement for system initialization complicates the portrait scanning process for more general poses. 

\subsection{3D Self-portrait Using Bundle Adjustment}
To suppress the accumulated error in incremental fusion, another branch of 3D self-portrait methods also utilizes bundle adjustment algorithms. 
Based on KinectFusion\cite{Newcombe2011KinectFusion}, Tong \textit{et al.} \cite{tong2012scanning} used 3 Kinects and a turntable for data capture and non-rigid bundle adjustment for portrait reconstruction. 
Cui \textit{et al.} \cite{cui2013kinectavatar} achieved self-rotating portrait reconstruction via non-rigid bundle. However, the efficiency is low due to large partial scan numbers. 
Wang \textit{et al.} \cite{wang2017templateless} conducted bundle adjustment for all point sets without volumetric fusion, which leads to over-smoothed results. 
The method in \cite{li20133d} is a very related work to ours for it also fuses large partial scans for portrait reconstruction. However, it needs the subject to keep static during the partial scanning process, thus cannot handle self-rotating reconstructions. 

\begin{figure*}[ht]
    \centering
    \includegraphics[width=\linewidth]{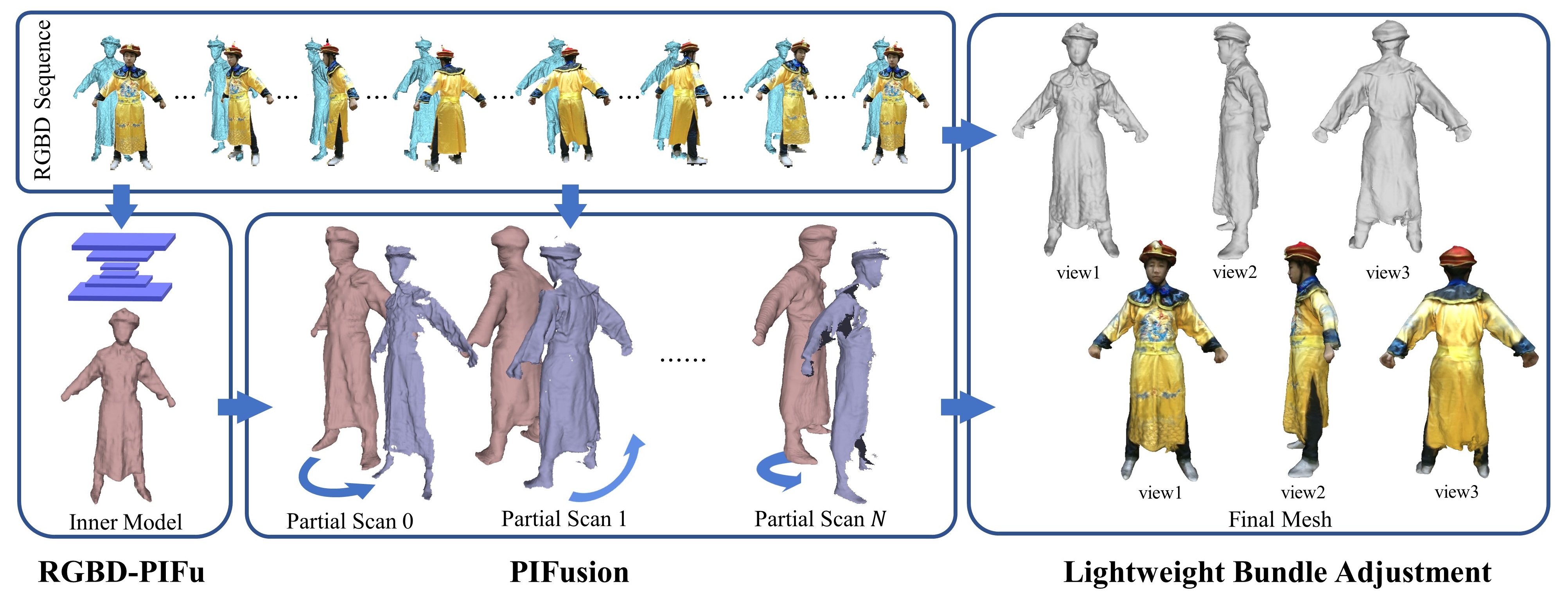}
    \caption{ System pipeline. In the first frame, we utilize RGBD-PIFu to generate a roughly correct inner model as a prior. 
    Then we perform PIFusion to generate large and accurate partial scans while the performer is turning around in front of the RGBD sensor. 
    Finally, we conduct lightweight bundle adjustment to merge all the partial scans and generate an accurate and detailed 3D portrait.}
    \label{fig:overview_system}
\end{figure*}

Besides above RGBD methods, using a RGB (without depth) video of a rotating human to reconstruct a plausible portrait is also a practical direction. Alldieck \textit{et al.} \cite{alldieck2018video, alldieck20183DV, alldieck19cvpr} used silhouette-based joint optimization and Zhu \textit{et al.} \cite{zhu2017videobased} used multi-view stereo technologies. However, current methods in this direction still rely on offsetting parametric models to represent cloth, which inherently limits their performance for more general clothed human reconstruction. Moreover, sparse feature points from RGB videos are not sufficient for detailed dense surface reconstruction.

%%% section4
\section{Overview}
\label{sec:overview}
As shown in Fig.~\ref{fig:overview_system}, given an RGBD sequence with a naturally self-rotating motion of the subject, our system performs 3 steps sequentially:
\begin{enumerate}[leftmargin=*]
	\setlength{\itemsep}{0pt}
	\setlength{\parsep}{0pt}
	\setlength{\parskip}{0pt}
	\item \textbf{RGBD-PIFu:} In this step, we use a neural network to infer a roughly accurate model of the subject from the first RGBD frame.
	\item \textbf{PIFusion:} For each frame, we first perform double-layer-based non-rigid tracking with the inferred model as the inner layer and then fuse the observations into the reference frame using the traditional non-rigid fusion method. Finally, non-rigid volumetric deformation is used to further optimize the inner model to improve both tracking and the fusion accuracy.
The partial scans are then generated by splitting the whole sequence into several chunks and fusing each chunk separately.
	\item \textbf{Lightweight bundle adjustment:} In each iteration, we first use key frame selection to select effective key frames to construct the live depth and silhouette terms. Then, joint optimization is performed to not only assemble all the partial scans in the reference frame but also optimize the warping fields to live key frames alternately.
\end{enumerate}

\begin{figure}[t]
	\centering
	\includegraphics[width=0.9\linewidth]{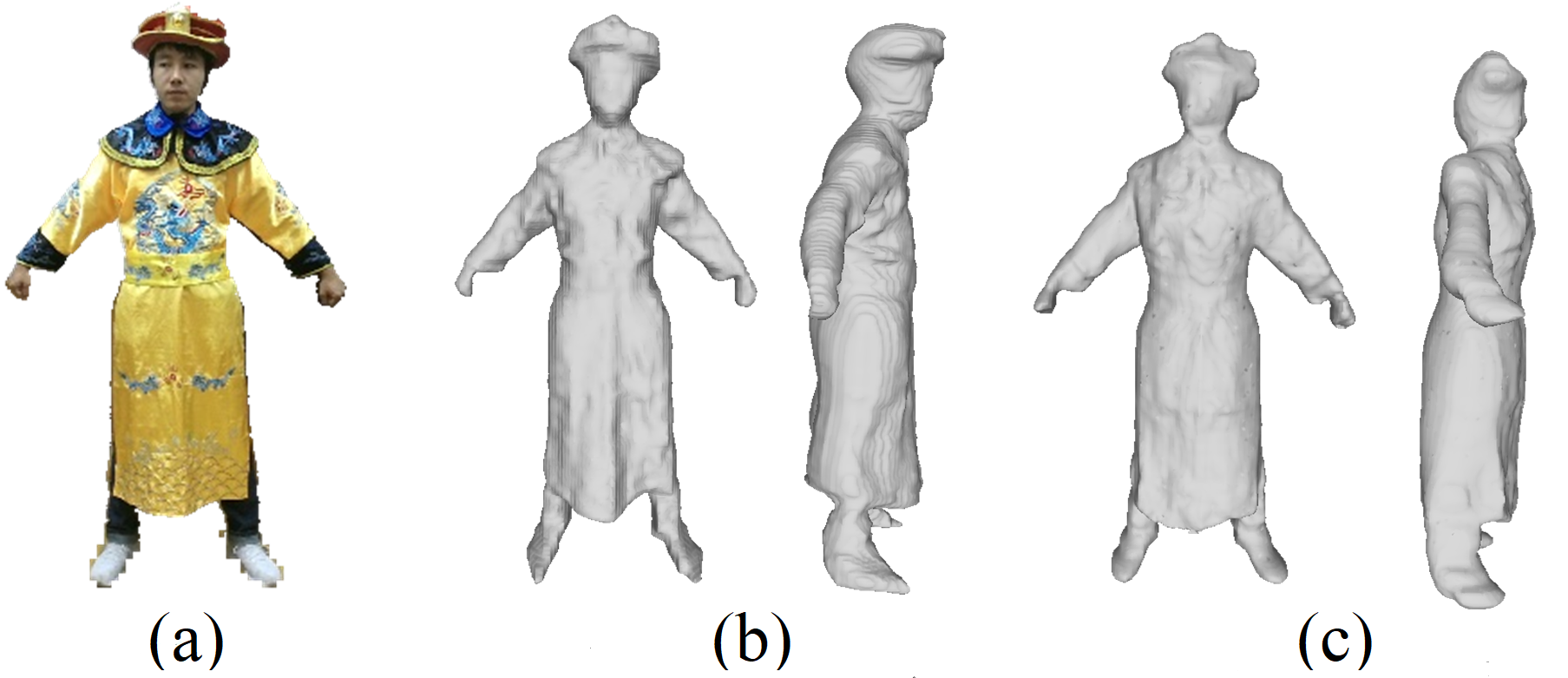}
\caption{Comparison of RGBD-PIFu and PIFu \cite{pifuSHNMKL19}. (a) Reference color image; (b) RGBD-PIFu result; (c) PIFu result.}
	\label{fig:comp of pifu}
\end{figure}
\section{RGBD-PIFu}
In this work, we extend pixel-aligned implicit functions (PIFu)\cite{pifuSHNMKL19} and propose RGBD-PIFu for 3D self-portrait inference from an RGBD image.
PIFu is a spatially aligned representation for 3D surfaces. It is a level-set function $f$ that defines the surface implicitly, e.g., $f(X)=0, X\in \mathbb{R}^3$. In our RGBD-PIFu method, this function is expressed as a composite function $f$, which consists of a fully convolutional RGBD image encoder $g$ and an implicit function $h$ represented by multilayer perceptrons:
\begin{equation}
f(X;I)=h(G(x;I),X_z),\, X\in \mathbb{R}^3,
\end{equation}
where $I$ is the input RGBD image, $x=\pi (X)$ is the 2D projection of a 3D point $X$, $G(x;I)$ is the feature vector of $x$ on the encoded feature map $g(I)$, and $X_z$ is the depth value of $X$. Different from \cite{pifuSHNMKL19}, our image encoder also encodes depth information, which forces the inner model to be consistent with the depth input, thus resolving the depth ambiguity problem and improving the reconstruction accuracy.
The training loss is defined as the mean squared error:
\begin{equation}
\mathcal{L}=\frac{1}{n} \sum_{i=1}^{n}\left|f\left(X_i;I\right)-f^{*}\left(X_{i}\right)\right|^{2},
\end{equation}
where $X_i$ is a sampled point, $f^*(X_i)$ is the ground-truth value, and $n$ is the number of sampled points.

In the model inference stage, to avoid dense sampling of the implicit function as in \cite{pifuSHNMKL19}, we utilize the depth input to ignore empty regions and only perform uniform sampling of the implicit function in the invisible regions. The isosurface is extracted by the marching cube algorithm \cite{Lorensen1987marching}. By incorporating depth features, our network is more robust and accurate than the original RGB-PIFu, thus producing a better mesh as the inner model for robust fusion performance, as shown in Fig.~\ref{fig:comp of pifu}.

\section{PIFusion}
\label{sec:PIFusion}
\subsection{Initialization}
\label{subsec:Init}
In the first frame, we initialize the TSDF (truncated signed distance function) volume by direct depth map projection and then fit the inner model to the initialized TSDF volume.
The deformation node graph (\cite{Sumner2007embedded}) is then uniformly sampled on the inner model using geodesic distance, which is used to parameterize the non-rigid deformation of the fused surface and the inner model.

\subsection{Double-layer Non-rigid Tracking}
\label{subsec:Tarcking}

Given the inner model and the fused mesh (i.e., the double-layer surface) in the $(t-1)$-th frame, we need to deform them to track the depth map in the $t$-th frame. Different from DynamicFusion \cite{newcombe2015dynamic}, an inner layer is used to assist non-rigid tracking. Hence, there are two types of correspondences: one is between the fused mesh (outer layer) and the depth observation, and the other is between the inner model (inner layer) and the depth observation.
The energy function is then formulated as:
\begin{equation}
\label{eq:tracking energy}
E_{\text{tracking}}=\lambda_{\text{outer}}E_{\text{outer}}+\lambda_{\text{inner}}E_{\text{inner}}+\lambda_{\text{smooth}}E_{\text{smooth}},
\end{equation}
where $E_{\text{outer}}$ and $E_{\text{inner}}$ are the energies of the two types of correspondences, $E_{\text{smooth}}$ is a smooth term to regularize local as-rigid-as-possible deformations, and $\lambda_{\text{outer}},\lambda_{\text{inner}},\lambda_{\text{smooth}}$ are the term weights.

\noindent{\bf Outer and Inner Term} The two terms measure the misalignment between the double layers and the depth map, and they have similar formulations:
\begin{equation}
\begin{split}
E_{\text{outer/inner}}=\sum_{(\mathbf{v},\mathbf{u})\in \mathcal{C}_{\text{outer/inner}}}\left| \mathbf{\hat{n}}_{\mathbf{v}}^\top (\mathbf{\hat{v}}-\mathbf{u}) \right|^2,
%\\
%E_{\text{inner}}=\sum_{(\mathbf{v},\mathbf{u})\in \mathcal{C}_2}\left| \mathbf{\hat{n}}_{\mathbf{v}}^\top (\mathbf{\hat{v}}-\mathbf{u}) \right|^2,
\end{split}
\end{equation}
where $\mathcal{C}_{\text{outer}}$ and $\mathcal{C}_{\text{inner}}$ are two types of correspondence sets, and $(\mathbf{v},\mathbf{u})$ is a correspondence pair; $\mathbf{v}$ is a vertex on the outer layer (fused mesh) or the inner layer (inner model), and $\mathbf{u}$ is the closest point to $\mathbf{v}$ on the depth map. Note that $\mathbf{v}$ is the coordinate in the reference frame, while $\mathbf{\hat{v}}$ and $\mathbf{\hat{n}}_{\mathbf{v}}$ are the position and normal of $\mathbf{v}$ in the live frame warped by its KNN nodes using dual quaternion blending:
\begin{equation}
\mathbf{T}(\mathbf{v})=SE3\left(\sum_{k\in\mathcal{N}(\mathbf{v})}w(k,\mathbf{v})\mathbf{dq}_k\right),
\end{equation}
where $\mathbf{dq}_k$ is the dual quaternion of the $k$-th node, $SE3(\cdot)$ maps a dual quaternion to the {\bf SE}(3) space, $\mathcal{N}(\mathbf{v})$ are the KNN nodes of $\mathbf{v}$, $w(k, \mathbf{v})=\exp (-\|\mathbf{v}-\mathbf{x}_{k}\|_{2}^{2} /(2 r^{2}))$ is the blending weight, $\mathbf{x}_k$ is the position of the $k$-th node, and $r$ is the active radius.

\noindent{\bf Smooth Term} The smooth term is defined on all edges of the node graph to guarantee local rigid deformation. This term is defined as
\begin{equation}
E_{\text{smooth}}=\sum_{i}\sum_{j\in\mathcal{N}(i)} \left\| \mathbf{T}_i \mathbf{x}_j - \mathbf{T}_j \mathbf{x}_j\right\|_2^2,
\label{eq:smooth term}
\end{equation}
where $\mathbf{T}_i$ and $\mathbf{T}_j$ are the transformations associated with the $i$-th and $j$-th nodes, and $\mathbf{x}_i$ and $\mathbf{x}_j$ are the positions of the $i$-th and $j$-th nodes in the reference frame, respectively.

We solve Eq.~\ref{eq:tracking energy} by the iterative closest point (ICP) algorithm and use the Gauss-Newton algorithm to solve the energy optimization problem. After tracking, we use the typical fusion method \cite{newcombe2015dynamic} to fuse the current depth observations and update the TSDF volume.

\subsection{Non-rigid Volumetric Deformation}
\label{subsec:volumetric opt}
The initial inner model inferred by RGBD-PIFu  is by no means accurate enough for double-layer surface tracking, and the correspondences between the inner model and the depth map may even reduce the tracking performance.
To deal with this issue, inspired by \cite{yu2018doublefusion}, we conduct a non-rigid volumetric deformation algorithm to continue correcting the inner model by fitting it to the fused mesh (i.e., the 0-level set of the TSDF) in the reference volume.
Moreover, the weight of the inner term, $\lambda_{\text{inner}}$ in Eq.~\ref{eq:tracking energy}, is also designed to decrease along the ICP iterations to enable a more accurate outer surface fitting performance.

We utilize the initialized node graph to parameterize the non-rigid deformation of the inner model. Given the updated TSDF volume of the fused mesh, the energy function of non-rigid volumetric deformation is defined as:

\begin{equation}
E_{\text{vol}}=E_{\text{tsdf}}+\lambda_{\text{smooth}}E_{\text{smooth}},
\end{equation}
where $E_{\text{tsdf}}$ measures the misalignment error between the inner model and the isosurface at threshold 0, and $E_{\text{smooth}}$ is the same as Eq.~\ref{eq:smooth term}. The TSDF term is defined as
\begin{equation}
E_{\text{tsdf}}=\sum_{\mathbf{v}\in\mathbf{T}}\left|\text{TSDF}(\hat{\mathbf{v}})\right|^2,
\end{equation}
where $\mathbf{T}$ is the initial inner model without non-rigid deformations in the reference frame, $\mathbf{v}$ is a vertex of $\mathbf{T}$, $\mathbf{\hat{v}}$ is the position warped by the KNN nodes of $\mathbf{v}$, TSDF($\cdot$) is a trilinear sampling function that takes a point in the reference frame and returns the interpolated TSDF value. By minimizing the squared sum of the TSDF values of all the vertices of the deformed inner model, the inner model will perfectly align with the fused mesh in the reference frame.

For the next frame, the corrected inner model is warped to the live frame to search for correspondences in the tracking step. This step provides more accurate correspondences and significantly improves the registration accuracy compared with directly warping the initial inner model.

\subsection{Partial Scan Fusion}
\label{subsec: fuse partial scans}
To guarantee that the following bundle adjustment is only conducted on a small number of partial scans, we fuse the partial scans within several large chunks of the whole sequence in the reference frame.
Specifically, given a sequence of the performer turning around in front of the sensor, we calculate the orientation of the performer and then split the whole sequence into 5 chunks, which cover the front, back and two side views of the performer.
Due to the accumulated error, the first and last partial scans that compose a loop may not align very well. The proposed lightweight bundle adjustment will resolve this problem and finally generate accurate 3D portraits.

\section{Lightweight Bundle Adjustment}
\label{sec:bundle adjustment}
\begin{figure}[t]
    \centering
    \includegraphics[width=\linewidth]{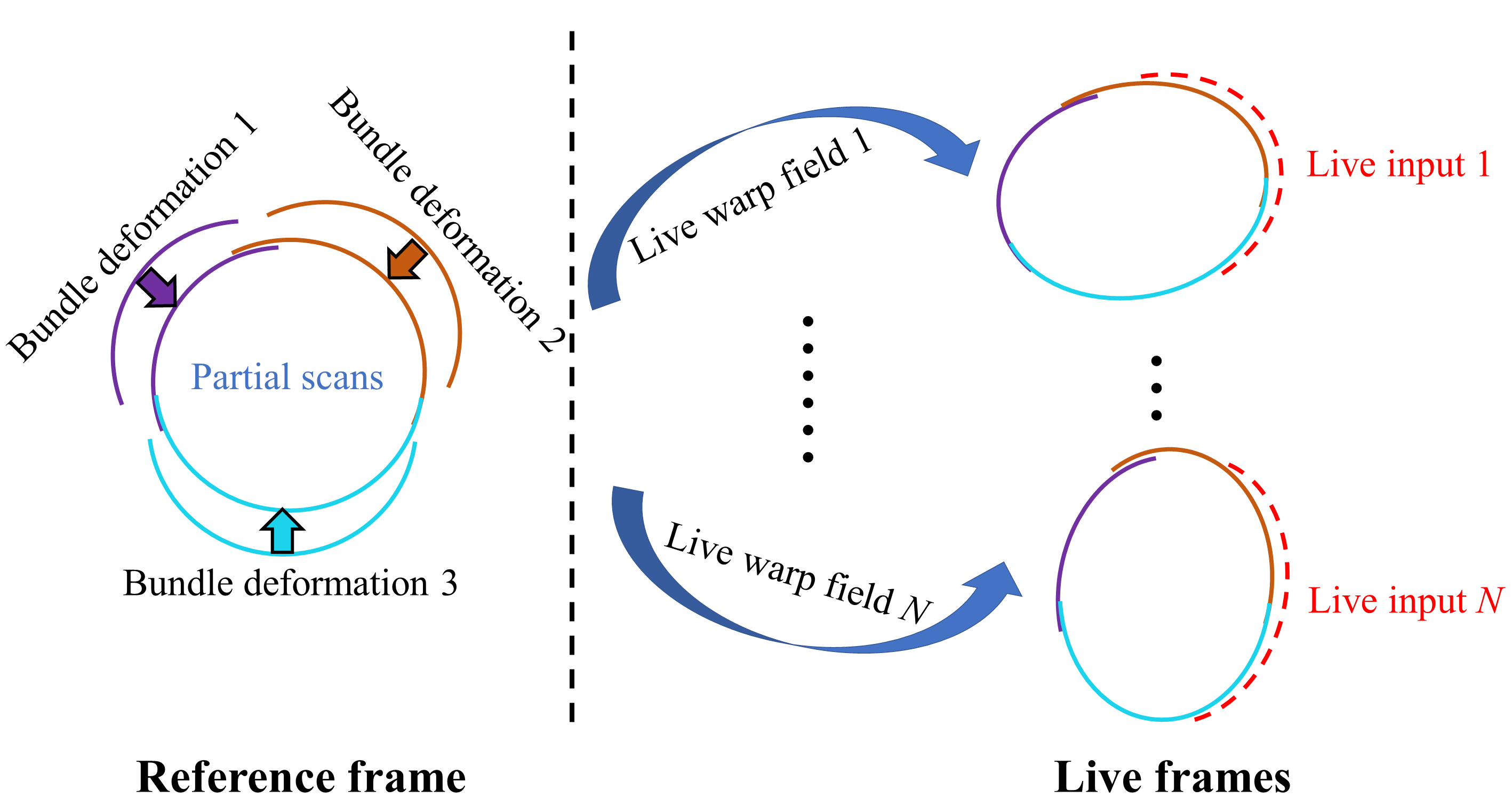}
\caption{Illustration of bundle adjustment with joint optimization. The bundle deformations are optimized to ``loop'' these partial scans in the reference frame, while the live warp fields are optimized to deform the partial scans to fit live input.}
    \label{fig:bundle_adjustment}
\end{figure}

Regarding non-rigid bundle adjustment (BA), we argue that a well-looped model after typical BA is an accurate model.
Our insight is that after BA, all the partial scans should not only construct a looped model in the reference frame but also be well fitted to all the live observations after non-rigid warping using the live warp fields.
To this end, we propose an efficient algorithm to jointly optimize the bundle deformations (for loop closure reconstruction) and live warp fields (for live depth fitting), as shown in Fig.~\ref{fig:bundle_adjustment}.
Novel energy terms, including the live depth and silhouette energies, are incorporated to enforce the consistency between the warped partial scans and live depth inputs.
However, optimizing the live warp fields corresponding to all the live frames in bundle adjustment will significantly decrease the efficiency. In practice, we found that performing live depth fitting on only several key frames is sufficient for generating accurate results. Therefore, we propose a key frame selection strategy to select effective key frames by sorting the live depth and silhouette energies.

\subsection{Joint Optimization}
After PIFusion, we can acquire $N$ partial scans.
As illustrated in Fig.~\ref{fig:bundle_adjustment}, we first construct a node graph for each partial scan for describing the bundle deformation, which is then optimized using loop closure correspondences to deform the partial scan to ``loop'' with the others in the reference frame.
Moreover, {all the partial scans are deformed together} to fit each live frame by optimizing the corresponding live warp field, which is similar to the non-rigid tracking in PIFusion.
As a result, each partial scan has its own bundle deformation, and all the partial scans share live warp fields in common.

We solve the joint optimization problem by optimizing the bundle deformations and live warp fields alternately. In each iteration, both bundle deformations and live warp fields will be updated to minimize the total energy.

\subsection{Key Frame Selection}
To maintain the efficiency of our algorithm, we propose a key frame selection strategy for constructing efficient and effective live depth fitting terms. Specifically, we uniformly divide the whole sequence into $K$ segments, and before each iteration of joint optimization, for each frame, we calculate two types of metrics: the geometric misalignment error and the silhouette error. The first metric is the misalignment between warped partial scans and the corresponding input depth point cloud.
The silhouette error is calculated by first rendering a mask map in the camera view using all the warped partial scans and then calculating the difference between the rendered mask and the input silhouette.
We then select the frames with the largest geometric misalignment error and silhouette error in each segment as depth key frames $\mathcal{K}_{\text{dep}}$ and silhouette key frames $\mathcal{K}_{\text{sil}}$, respectively.

\subsection{Formulation}
Different from other bundle adjustment algorithms (\cite{dou20153d, wang2017templateless}), we not only ``loop''' these partial scans but also introduce live frame observations into the optimization procedure to improve the accuracy. The total energy function is defined as
\begin{equation}
\begin{split}
E(\mathbf{W}_b^j,\mathbf{W}_l^i)=\lambda_{\text{loop}}E_{\text{loop}}+\lambda_{\text{depth}}E_{\text{depth}}\\+\lambda_{\text{silhouette}}E_{\text{silhouette}}+\lambda_{\text{smooth}}E_{\text{smooth}},
\label{eq:bundle}
\end{split}
\end{equation}
where $\mathbf{W}_b^j$ is the bundle deformation corresponding to the $j$-th partial scan, $\mathbf{W}_l^i$ is the live warp field from the reference frame to the $i$-th key frame, and $E_{\text{loop}}, E_{\text{depth}}, E_{\text{silhouette}}$ and $E_{\text{smooth}}$ are the energies of loop closure, live depth, live silhouette and smooth regularization terms, respectively.

In each iteration, we optimize the bundle deformation and live warp fields alternately to minimize Eq.~\ref{eq:bundle}. Note that after PIFusion, although the partial scans have already been well aligned with the live depth inputs, the live warp fields are still not accurate enough to guarantee all the fused partial scans to construct a loop in the reference frame directly. Thus, the bundle deformation in the reference frame will conflict with live depth fitting in the live frames without simultaneously optimizing the live warp fields.

\noindent {\bf Loop Term} The loop term measures the amount of misalignment among these partial scans and is defined as
\begin{equation}
E_{\text{loop}}=\sum_{i,j=1\atop i\neq j}^N \sum_{(\mathbf{v}_p,\mathbf{v}_q)\in \mathcal{C}_{i,j}}\left|\mathbf{W}_b^i(\mathbf{n}_p)^\top\left(\mathbf{W}_b^i(\mathbf{v}_p)-\mathbf{W}_b^j(\mathbf{v}_q)\right)\right|^2,
\end{equation}
where $N$ is the number of partial scans, $\mathcal{C}_{i,j}$ is the correspondence set between the $i$-th and $j$-th partial scans acquired by searching the closest points, $(\mathbf{v}_p,\mathbf{v}_q)$ is a correspondence pair, $\mathbf{v}_p$ and $\mathbf{v}_q$ are vertices on the $i$-th and $j$-th partial scans, respectively, $\mathbf{n}_p$ is the normal of $\mathbf{v}_p$ on the $i$-th partial scan, and $\mathbf{W}_b^i(\mathbf{v}_p)$ and $\mathbf{W}_b^i(\mathbf{n}_p)$ represent the position and normal warped by bundle deformation. This term forces all the partial scans to register with each other in the reference frame.

\noindent{\bf Live Depth Term} This term measures the misalignment of all the partial scans with all the depth maps in $\mathcal{K}_{\text{depth}}$:
\begin{equation}
\begin{split}
E_{\text{depth}}=&\sum_{i=1}^K \sum_{j=1}^N \sum_{(\mathbf{v},\mathbf{u})\in \mathcal{D}_{j,i}} \\& \left| \mathbf{W}_l^i(\mathbf{W}_b^j(\mathbf{n}))^\top \left( \mathbf{W}_l^i(\mathbf{W}_b^j(\mathbf{v}))-\mathbf{u} \right) \right|^2,
\end{split}
\end{equation}
where $K=|\mathcal{K}_{\text{depth}}|$ is the number of key frames, $\mathcal{D}_{j,i}$ is the correspondence set between the $j$-th partial scan and the depth map in the $i$-th key frame, $(\mathbf{v},\mathbf{u})$ is a correspondence pair, $\mathbf{v}$ is a vertex on the $j$-th partial scan, $\mathbf{u}$ is a point on the depth map, and $\mathbf{W}_l^i(\cdot)$ takes a point or normal in the reference frame as input and returns the warped position or normal in the $i$-th key frame. This term is designed to force partial scans to align with depth point clouds in $\mathcal{K}_{\text{depth}}$.

\noindent{\bf Live Silhouette Term} This term measures the misalignment between the rendered mask of all the warped partial scans and the input mask of the body shape in the live frames.
Similar to LiveCap\cite{habermann2019livecap}, we preprocess the input mask using the distance transform. For the $i$-th key frame, we render a mask image of all partial scans deformed by the live warp field and then filter a boundary vertex set $\mathcal{B}_i$. We define the live silhouette term as
\begin{equation}
E_{\text{silhouette}}=\sum_{i=1}^K\sum_{j=1}^N \sum_{\mathbf{v}_j\in\mathcal{B}_i} d_j\cdot \left|I_{\text{DT}}^i(\pi(\mathbf{W}_l^i(\mathbf{W}_b^j(\mathbf{v}_j))))\right|^2,
\end{equation}
where $K=|\mathcal{K}_{\text{sil}}|$ is the number of key frames, $\mathbf{v}_j$ is a boundary vertex on the $j$-th partial scan (note that the boundary means that the vertex is projected near the boundaries of the rendered mask image rather than the boundary of this partial scan), $d_j\in\{-1,+1\}$ is an indicated value that indicates the correct direction in the distance field\cite{habermann2019livecap}, $I_{\text{DT}}^i$ is the distance-transformed image of the input mask and $\pi(\cdot)$ is the projection function. This term will deform the shape of the partial scans to fit with the input silhouettes.

The smooth term is defined similarly to Eq.~\ref{eq:smooth term}. We solve Eq.~\ref{eq:bundle} using the Gauss-Newton method. Within each iteration, we construct a large sparse system of linear equations and then utilize an efficient preconditioned conjugate gradient (PCG) solver on a GPU to obtain the updates.

\subsection{Non-rigid Multi-texturing}
After lightweight bundle adjustment, we fuse all the partial scans into a watertight mesh using Poisson reconstruction \cite{Kazhdan:2006:PSR:1281957.1281965}. For each live frame, we project each visible vertex to the color image to retrieve a color value. After processing all the frames, we blend the retrieved color values according to the normal direction and obtain the final vertex color. Specifically, for vertex $\mathbf{v}_i$, we calculate its color $C_{\mathbf{v}_i}$ as a weighted average of the color values retrieved from all the live frames.
The blending weight $\omega_{i,j}$ is defined as:
\begin{equation}
    \omega_{i,j}=\left\{\begin{matrix}
0 &,\text{$\mathbf{v}_i$ is invisible in the $j$-th frame}\\ 
\frac{|\mathbf{n}_{\mathbf{v}_i}\cdot\hat{z}|}{|\mathbf{n}_{\mathbf{v}_i}|} &,\text{$\mathbf{v}_i$ is visible in the $j$-th frame}
\end{matrix}\right.,
\end{equation}
where $\mathbf{n}_{\mathbf{v}_i}$ is the normal of $\mathbf{v}_i$ and $\hat{z}$ is the direction the camera is looking. To avoid oversmoothing, for each vertex, only the top $15\%$ of  weighted color values are blended.

\begin{figure*}[t]
    \centering
    \includegraphics[width=\linewidth]{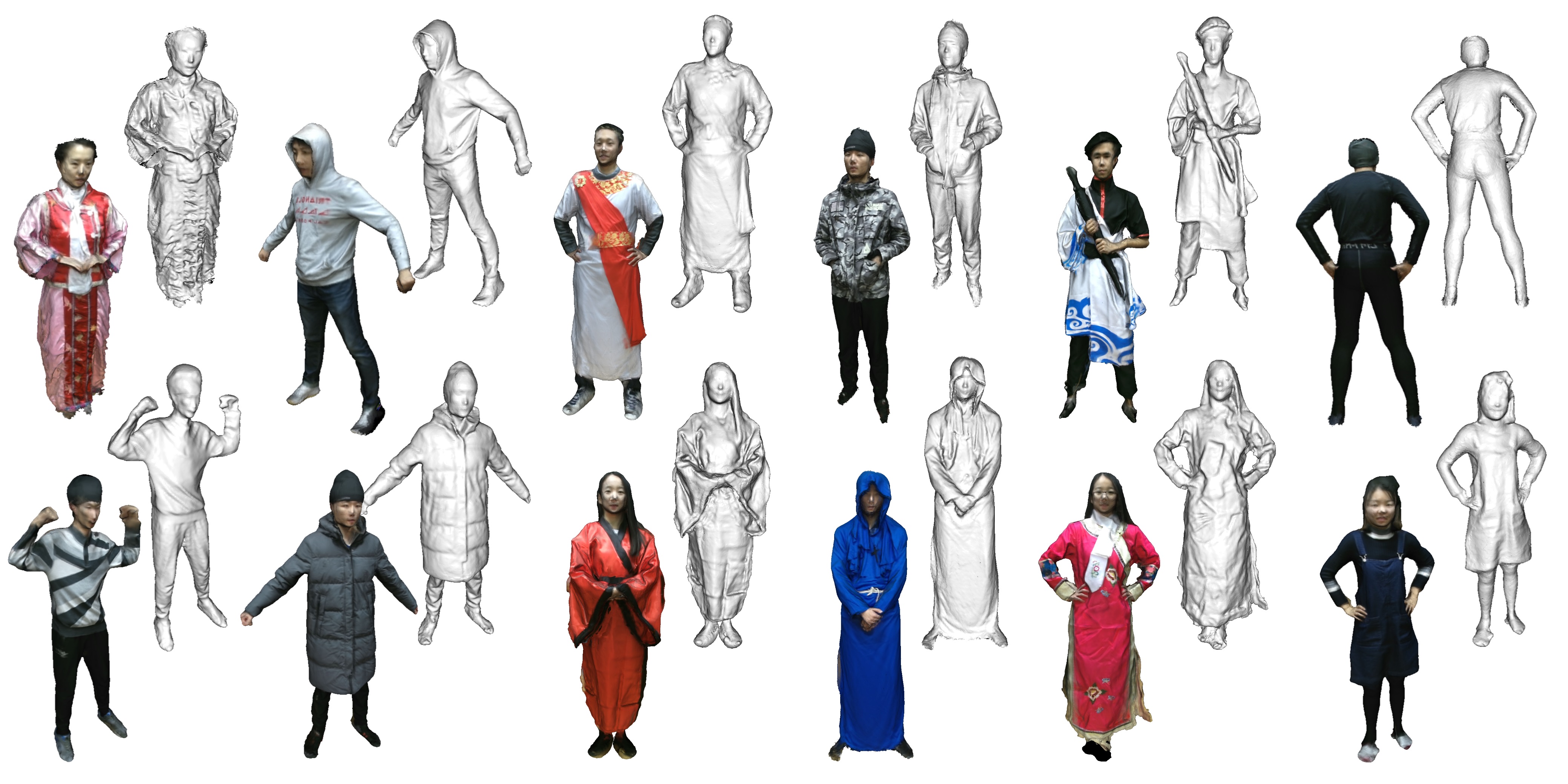}
\caption{Examples of 3D portraits acquired by our system.}
    \label{fig:results}
\end{figure*}

\section{Results}
\label{sec:result}
In this section, we first report the system performance and our implementation. Then, we compare our method with current state-of-the-art works. Finally, we evaluate the core parts of our system. In Fig.~\ref{fig:results}, we demonstrate several 3D portraits acquired by our system.
\subsection{Performance and Implementation}
\label{subsec:implementation}
Our 3D self-portrait system is very efficient. The whole pipeline is implemented on one NVIDIA Geforce RTX 2080Ti GPU. The initialization that contains generating the inner model by RGBD-PIFu and initializing PIFusion takes almost 10 seconds. PIFusion runs in real-time (at 30 ms per frame). For each frame, the tracking, volumetric deformation and fusion take 20 ms, 3 ms and 6 ms, respectively.

Similar to \cite{pifuSHNMKL19}, we adapt a stacked hourglass network \cite{hourglass} as the image encoder, and the implicit function is presented by a MLP with 257, 1024, 512, 256, 128, and 1 neurons in each layer. We render the Twindom dataset (\url{https://web.twindom.com/}) to acquire depth and color images and utilize 3500 images to train this network. When training, the batch size is 4, the learning rate is $1\times 10^{-3}$, and the number of epochs is 28. The training procedure takes one day on one RTX 2080Ti GPU.

In the tracking of PIFusion, the number of ICP iterations is 5 per frame, and we set $\lambda_{\text{outer}}=1.0$, $\lambda_{\text{inner}}=1.0$, and $\lambda_{\text{smooth}}=5.0$, while $\lambda_{\text{inner}}$ will decrease linearly as the iteration continues. For each vertex, we use its 4 nearest neighbors for non-rigid deformation, and the number of neighbors of each node is 8. In bundle adjustment, the numbers of partial scans and key frames are both 5, we set $\lambda_{\text{loop}}=1.0$, $\lambda_{\text{depth}}=0.5$, $\lambda_{\text{silhouette}}=0.001$ and $\lambda_{\text{smooth}}=2.0$, and the number of iterations is 25. This procedure takes only 15 seconds, and texturing takes 1 second.
\begin{figure}[t]
	\centering
	\includegraphics[width=\linewidth]{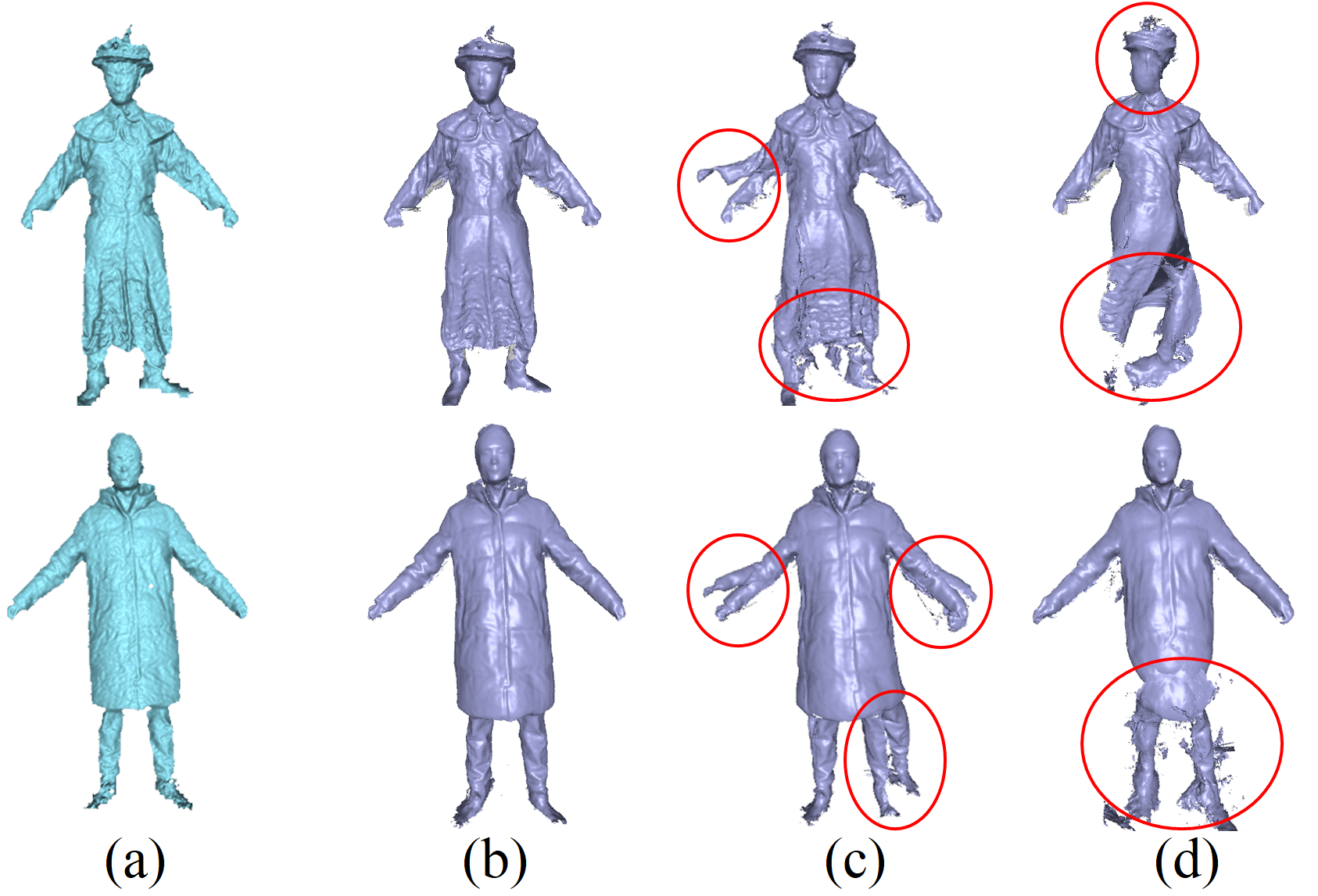}
\caption{Comparison of the proposed PIFusion, DynamicFusion \cite{newcombe2015dynamic} and DoubleFusion \cite{yu2018doublefusion} methods. (a) Reference depth input; (b), (c) and (d) are the results of PIFusion, DynamicFusion and DoubleFusion, respectively.}
	\label{fig:comp of fusion}
\end{figure}
\begin{figure}[t]
	\centering
	\includegraphics[width=\linewidth]{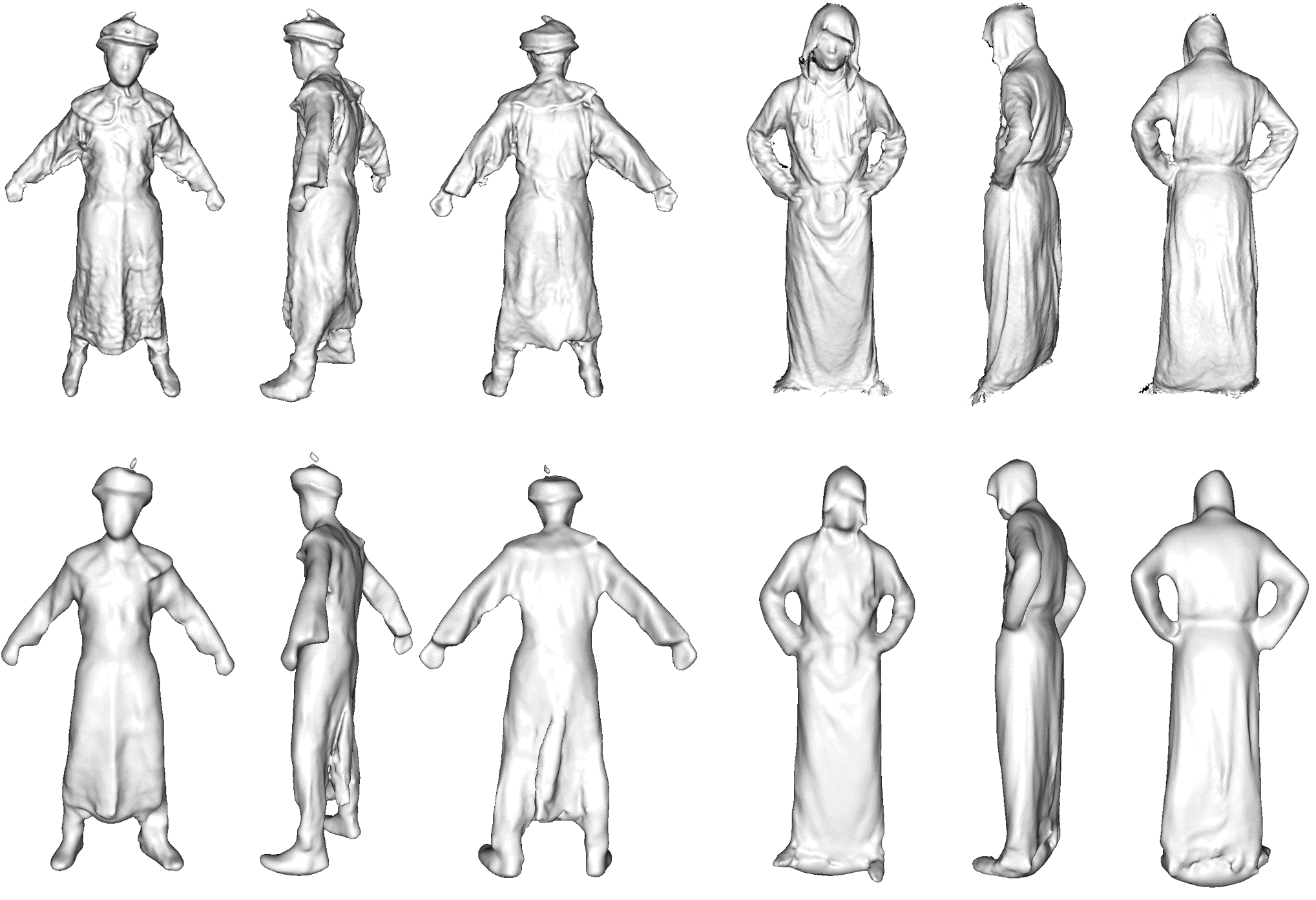}
\caption{Comparison of our method (the top row) with the method proposed by Wang \textit{et al.} \cite{wang2017templateless} (the bottom row).}
	\label{fig:comp with wkk}
\end{figure}

\subsection{Comparison}
\label{subsec:comparison}

\noindent\textbf{Comparison with Fusion Methods} We compared our fusion method PIFusion with DynamicFusion \cite{newcombe2015dynamic} and DoubleFusion \cite{yu2018doublefusion} using sequences captured by a Kinect V2 sensor. Fig.~\ref{fig:comp of fusion} demonstrates that our method improves the tracking and implicit loop-closure performance compared to the other methods, especially for subjects wearing loose clothes. Note that for this experiment, we use PIFusion to fuse the whole sequence without generating partial scans.

\noindent\textbf{Comparison with Bundle Adjustment Methods} We compare our method with the state-of-the-art non-rigid bundle adjustment method Wang \textit{et al.} \cite{wang2017templateless}. As shown in Fig.~\ref{fig:comp with wkk}, our method achieves much more detailed and accurate 3D self-portraits than \cite{wang2017templateless}. Moreover, as mentioned in the related work section of \cite{wang2017templateless}, although the results are plausible, \cite{li20133d} requires the subject to remain static several times during scanning, thus complicating the scanning process.

\subsection{Evaluation}
\label{subsec:evaluation}
\begin{figure}[t]
    \centering
    \includegraphics[width=\linewidth]{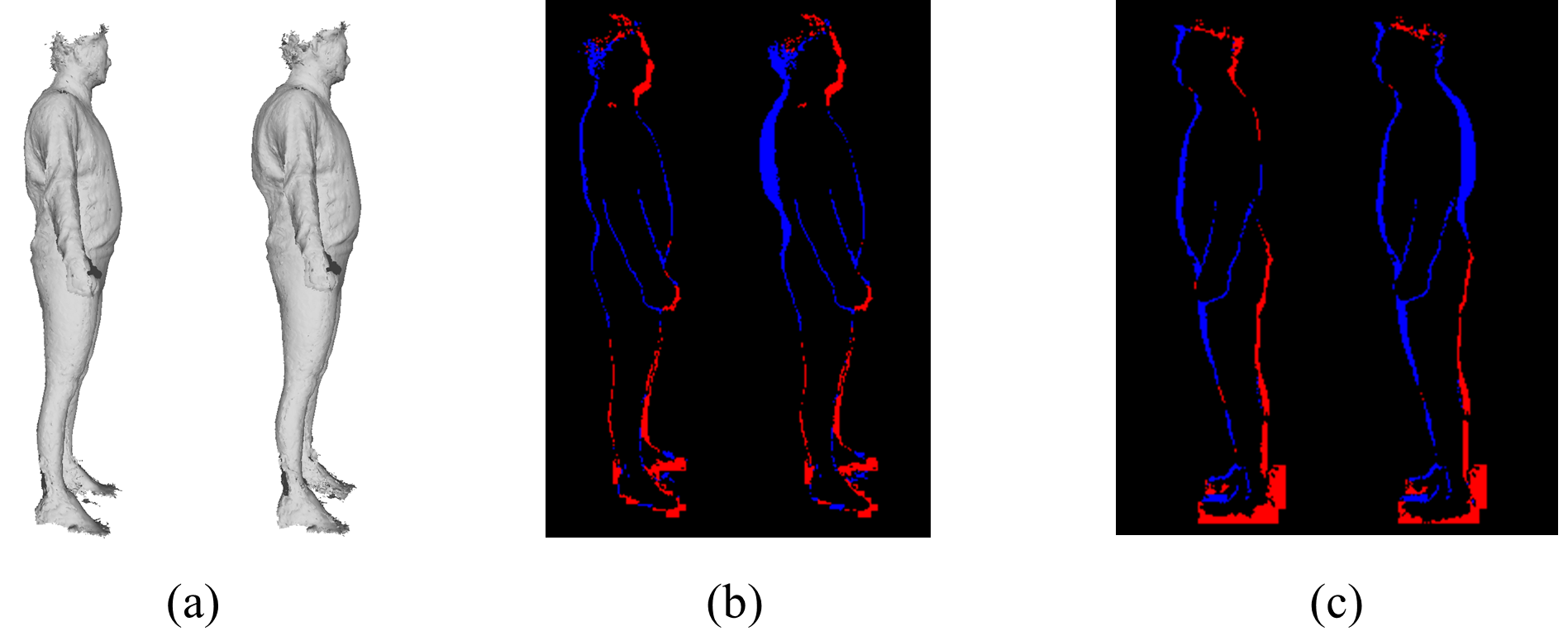}
\caption{Evaluation of the live silhouette term. (a) Optimized partial scans with (left) and without (right) the live silhouette term, (b) the mask error maps in the 1st key frame with (left) and without (right) the live silhouette term, (c) mask error maps in the 3rd key frame with (left) and without (right) the live silhouette term (nonblack pixels represent errors).}
    \label{fig:eval sil term}
\end{figure}

\begin{figure}[t]
	\centering
	\includegraphics[width=0.8\linewidth]{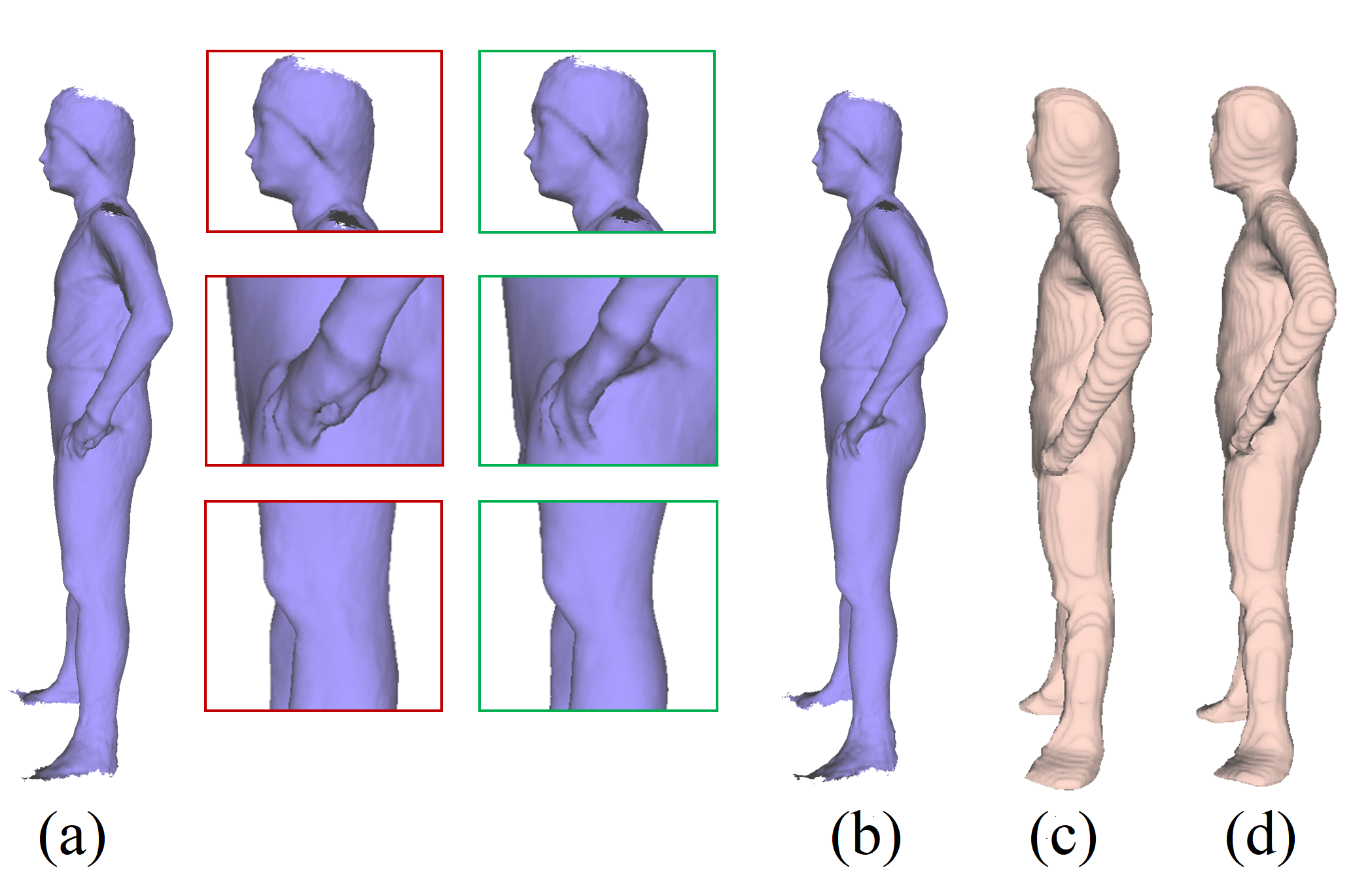}
\caption{Evaluation of the non-rigid volumetric deformation step. Fused mesh without (a) and with (b) volumetric deformation; (c) the original inner model generated by RGBD-PIFu; (d) the inner model after volumetric deformation.}
	\label{fig:eval of vol opt}
\end{figure}

\noindent{\bf Ablation Studies on Energy Terms}\\
\noindent{\bf -- Inner and Outer Terms in PIFusion} Without the inner term, PIFusion will degenerate into DynamicFusion \cite{newcombe2015dynamic}, which suffers from heavy drifts and tracking errors (Fig.~\ref{fig:comp of fusion}). Moreover, the lack of the outer term makes the final reconstruction accuracy fully depend on the accuracy of the shape prior, which is usually not accurate enough.\\
\noindent{\bf -- Live Silhouette Term in Bundle Adjustment} Fig.~\ref{fig:eval sil term} demonstrates that the live silhouette term could deform partial scans to be consistent with input silhouettes, thus further improving the accuracy of the optimized partial scans.\\
\noindent{\bf Non-rigid Volumetric Deformation} We evaluate the non-rigid volumetric deformation qualitatively, as shown in Fig.~\ref{fig:eval of vol opt}. The results demonstrate that without volumetric deformation, the fused geometry will highly depend on the original inner model generated by RGBD-PIFu.
With non-rigid volumetric deformation, the outer observations are introduced to update the inner model in each frame. This step will alleviate the errors brought by the inner model and improve the accuracy of our reconstruction results.

\begin{figure}[t]
	\centering
	\includegraphics[width=0.75\linewidth]{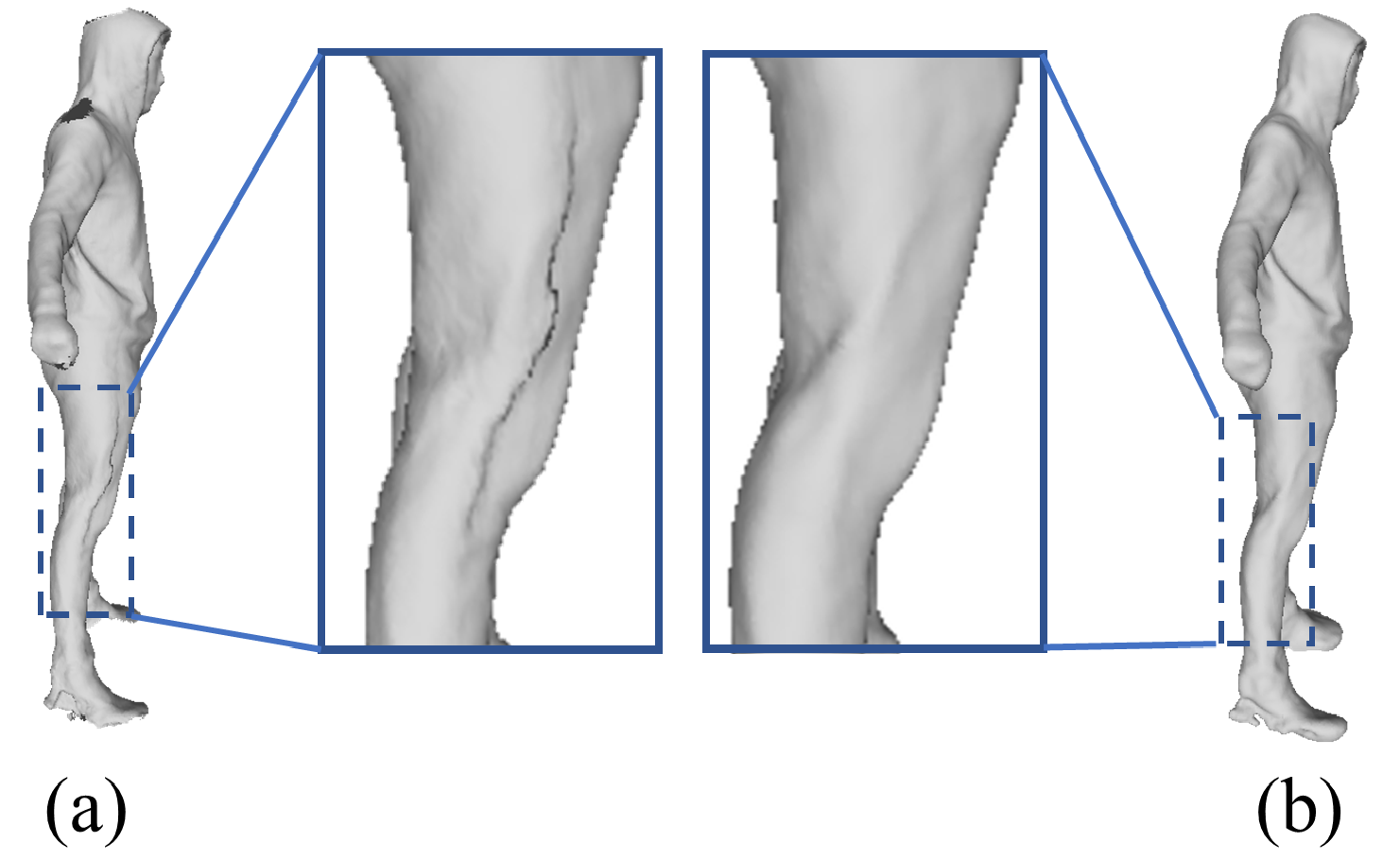}
\caption{Evaluation of the loop closure. (a) Fused mesh by PIFusion; (b) the reconstructed mesh after bundle adjustment.}
	\label{fig:eval of bundle with pifusion}
\end{figure}

\noindent{\bf Loop Closure} We compare the reconstructed result after lightweight bundle adjustment with the mesh completely fused using PIFusion in Fig.~\ref{fig:eval of bundle with pifusion}.
This comparison demonstrates that PIFusion still suffers from loop-closure problems, especially in the cases of challenging motion (very articulated motion) and an inaccurate initial inner model.
With the help of bundle adjustment, we can obtain more accurate portraits efficiently than the fusion methods.

\begin{figure}[t]
    \centering
    \includegraphics[width=\linewidth]{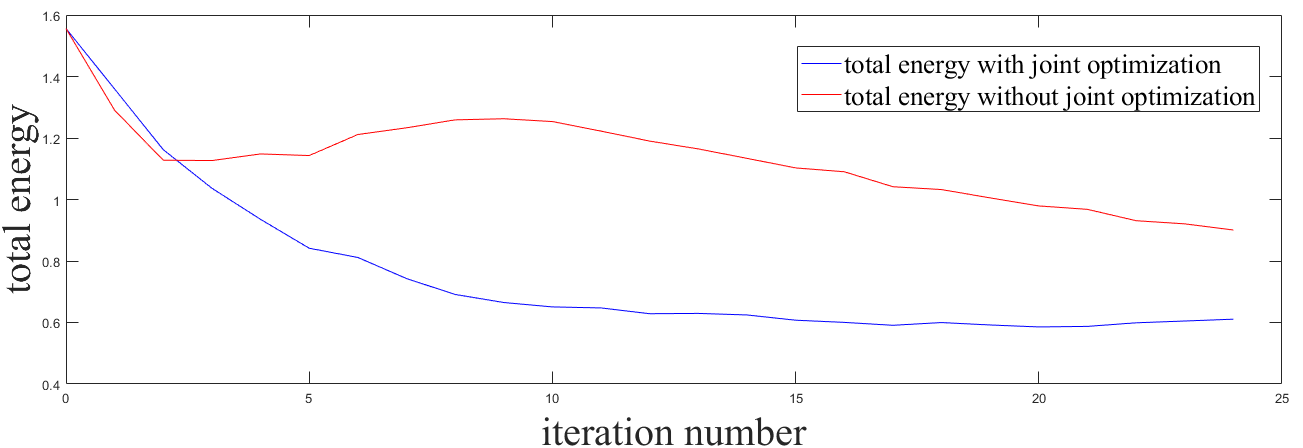}
\caption{The total energy in each bundle adjustment iteration.}
   \label{fig:eval alter}
\end{figure}

\noindent{\bf Joint Optimization} We evaluate the joint optimization by the total energy of Eq.~\ref{eq:bundle} in each iteration. 
{Fig.~\ref{fig:eval alter} demonstrates that joint optimization of both the bundle deformation and live warp fields can achieve a lower optimum than the method without joint optimization.}

\noindent {\bf Body Measurement} We quantitatively evaluated the accuracy of our results on body measurements. To evaluate the measurement error, we first utilized a laser scanner to obtain the ground-truth shape of a tight-clothed subject and then scanned the subject again using our system. Tab.~\ref{tab:body measurement} shows the measurement results of several body parts, which illustrates that our method acquires more accurate results than the state-of-the-art fusion-based body reconstruction method, DoubleFusion\cite{yu2018doublefusion}. Moreover, the proposed lightweight bundle adjustment method is effective in further improving the accuracy of the final reconstruction.

\begin{table}[t]
\centering
\begin{tabular}{|l|c|c|c|}
\hline
Method            & chest & waist & right knee \\ \hline
DoubleFusion\cite{yu2018doublefusion}      & 98.7  & 92.5  & 43.1       \\ \hline
PIFusion          & 97.6  & 87.2  & 40.7       \\ \hline
Bundle Adjustment & 94.6  & 84.5  & 39.6       \\ \hline
Ground Truth      & 91.2  & 79.7  & 37.7       \\ \hline
\end{tabular}
\caption{Evaluation of body measurements on case ``lz": the circumference of some body parts (cm).}
\label{tab:body measurement}
\end{table}

\section{Discussion}
\label{sec:discussion}

\noindent\textbf{Conclusion} In this paper, we have proposed a new method for robust and efficient 3D self-portrait reconstruction from a single RGBD camera.
We propose PIFusion, a novel volumetric non-rigid fusion method constrained by a learned shape prior, for generating large and accurate partial scans.
More importantly, the proposed lightweight bundle adjustment method not only guarantees the generation of a looped model in the reference frame but also ensures the alignment with live key observations, which further improves the accuracy of the final portrait without losing efficiency.
In conclusion, with the proposed method, users can conveniently obtain detailed and accurate 3D self-portraits in seconds.

\noindent\textbf{Limitations} Our method still relies on the completeness of the shape prior provided by RGBD-PIFu. Specifically, if the inferred shape prior loses some body parts, the final reconstruction may also lose these parts. Moreover, if some cases (e.g., object interactions) are not included in the training dataset of RGBD-PIFu, they may also not be well handled. However, growing the deformation node graph according to live observations may solve these problems.

\noindent\textbf{Acknowledgments} This paper is supported by the NSFC No.61827805, No.61531014 and No.61861166002.

\clearpage
{\small
\bibliographystyle{ieee}
\bibliography{egbib}
}

\end{document}